\documentclass[a4paper]{article}
\usepackage{INTERSPEECH2014,amssymb,amsmath,epsfig,framed, booktabs}
\usepackage{multirow}
\ninept
\def\reg{{\rm\ooalign{\hfil
     \raise.07ex\hbox{\scriptsize R}\hfil\crcr\mathhexbox20D}}}

\title{GMM-Free Flat Start Sequence-Discriminative DNN Training
}

\makeatletter
\def\name#1{\gdef\@name{#1\\}}
\makeatother\name{{\em G\'abor Gosztolya~$^{1,2}$, Tam\'as
Gr\'osz~$^2$, L\'aszl\'o T\'oth~$^1$}}

\address{$^1$ MTA-SZTE Research Group on Artificial Intelligence, Szeged, Hungary\\
$^2$ Department of Informatics, University of Szeged, Hungary\\
{\small \tt \{ ggabor, groszt, tothl \} @ inf.u-szeged.hu}}
\begin{document}
\maketitle
\begin{abstract}
Recently, attempts have been made to remove Gaussian mixture models
(GMM) from the training process of deep neural network-based hidden
Markov models (HMM/DNN). For the GMM-free training of a HMM/DNN
hybrid we have to solve two problems, namely the initial alignment
of the frame-level state labels and the creation of
context-dependent states. Although flat-start training via
iteratively realigning and retraining the DNN using a frame-level
error function is viable, it is quite cumbersome. Here, we propose
to use a sequence-discriminative training criterion for flat start.
While sequence-discriminative training is routinely applied only in
the final phase of model training, we show that with proper caution
it is also suitable for getting an alignment of context-independent
DNN models. For the construction of tied states we apply a recently
proposed KL-divergence-based state clustering method, hence our
whole training process is GMM-free. In the experimental evaluation
we found that the sequence-discriminative flat start training method
is not only significantly faster than the straightforward approach
of iterative retraining and realignment, but the word error rates
attained are slightly better as well.
\end{abstract}
\noindent{\bf Index Terms}: deep neural networks, flat start,
sequence discriminative DNN training

\section{Introduction}

While deep neural network (DNN) based speech recognizers have
recently replaced Gaussian mixture (GMM) based systems as the
state-of-the-art in ASR, the training process of HMM/DNN hybrids
still relies on the HMM/GMM framework. Conventionally, we start the
training of a HMM/DNN by constructing a HMM/GMM system, which is
then applied to get an alignment for the frame-level state labels.
These labels are then used as the training targets for the DNN. The
second task that requires GMMs is the state tying algorithm utilized
for the construction of context-dependent (CD) phone models. We
proposed a GMM-free solution for state clustering
earlier~\cite{gosztolya2015building}, and in this study we will
focus on the issue of obtaining the initial state alignment for
training the DNN.

The most convenient way of training the DNN component of a HMM/DNN
hybrid is by applying a frame-level error criterion, which is
usually the cross-entropy (CE) function. This solution, however,
requires frame-aligned training labels, while the training dataset
contains just orthographic transcripts in most cases. Trivially, one
may train a HMM/GMM system to get aligned labels, but this is
clearly a waste of resources.

The procedure for training HMM/GMM systems without alignment
information is commonly known as 'flat start
training'~\cite{htkbook}. This consists of initializing all phone
models with the same parameters, which would result in a uniform
alignment of phone boundaries in the first iteration of Baum-Welch
training. It is possible to construct a flat start-like training
procedure for CE-trained DNNs as well, by iteratively training and
realigning the DNN. For example, Senior et al. randomly initialized
their neural network~\cite{senior2014gmmfreeacoustic}, while Zhang
et al. trained their first model on equal-sized segments for each
state~\cite{zhang2014standalone}. As these solutions have a slow
convergence rate, they require a lot of training-realignment loops.

Although training the DNN at the frame level is straightforward, it
is clearly not optimal, as the recognition is performed and
evaluated at the sentence level. Within the framework of HMM/GMM
systems, several sequence-discrimina\-tive training methods have
been developed, and
these have now been adapted 
to HMM/DNN hybrids as
well~\cite{kingsbury2009latticebased,Vesely-str,grosz2014asequence}.
However, most authors apply sequence-discriminative criteria only in
the final phase of training, for the refinement of the DNN model.
That is, the first step is always CE-based training, either to
initialize the DNN
(e.g.~\cite{zhou2014sequence,mcdermott2014asynchronous,mohamed2010investigation})
or just to provide frame-level state labels
(e.g.~\cite{kingsbury2009latticebased,Vesely-str,saon2014comparison,wiesler2015investigations,chen2014joint}).

The Connectionist Temporal Classification (CTC) approach has
recently become very popular for training DNNs without an initial
time alignment being available~\cite{GravesICASSP13}. Rao et al.
proposed a flat start training procedure which is built on
CTC~\cite{rao2016flatstart}. However, CTC has several drawbacks
compared to MMI. First, it introduces blank labels, which require
special care in the later steps (e.g. CD state tying) of the
training process. Second, the CTC algorithm is not a
sequence-discriminative training method, so for the best performance
it has to be combined with techniques like sMBR
training~\cite{GravesICASSP13,rao2016flatstart}.



In contrast with the previous authors, here we propose a training
procedure that applies sequence-discriminative training in the
flat-start training phase. This requires several small modifications
compared to the standard usage of sequence-discriminative training,
which will be discussed in detail. In the experimental part we
compare the proposed method with the CE-based iterative
retraining-realignment procedure of Zhang et
al.~\cite{zhang2014standalone}, and we find that our method is
faster and gives slightly lower word error rates. Furthermore, we
can combine sequence-discriminative flat start training with the
Kullback-Leibler divergence-based state clustering method we
proposed recently~\cite{gosztolya2015building}. With this, we
eliminate all dependencies from a HMM/GMM system, making the whole
training procedure of context-dependent HMM/DNNs GMM-free.

\section{Flat-start training of HMM/DNN}
\label{section_flatstart}

Conventionally, the training of a HMM/DNN system is initiated by
training a HMM/GMM just to get time-aligned training labels.
Here, we compare two approaches that seek to eliminate GMMs from
this process. As the baseline method, we apply a simple solution
that iterates the loop of CE DNN training and realignment.
Afterwards, we propose an approach that creates time-aligned
transcriptions for the training data by training a DNN with a
sequence training criterion. From the wide variety of sequence
training methods, we opted for MMI (Maximum Mutual Information)
training~\cite{kingsbury2009latticebased}. Applying sequence
training to flat start requires some slight modifications, which we
will now discuss.

\subsection{Iterative CE Training and Realignment}

For comparison we will also test what is perhaps the most
straightforward solution for flat start DNN training, namely just
using the CE training criterion and iterating DNN training and
realignment. Here, we used the following algorithm that was based on
the description of Zhang et al.~\cite{zhang2014standalone}:
\begin{enumerate}
\item Train a DNN using uniformly segmented sound files.
\item Use the current DNN to realign the labels.
\item Train a randomly initialized DNN using the new
alignments.
\item Repeat steps 2--3 several times.
\end{enumerate}
The final DNN was utilized to create time-aligned labels for the
training set.

The main advantage of this method is that it requires only an
implementation of CE training for the DNN, and the realignment step
can also be readily performed by using standard ASR toolkits. The
drawback is that the procedure of retraining and realignment tends
to be rather time-consuming, which was also confirmed by our
experiments (see Section~\ref{sec_results}).

\subsection{Sequence-Discriminative Training Using MMI}

%

Several sequence-discriminative training criteria have been
developed for HMM/GMMs~\cite{discriminativebook} -- and adapted to
HMM/DNNs~\cite{kingsbury2009latticebased,
Vesely-str,wiesler2015investigations,Dong-seq} -- from which the
maximum mutual information (MMI) criterion is the oldest and
simplest. The MMI function measures the mutual information between
the distribution of the observation and the phoneme sequence.
Denoting the sequence of all observations by $O_u =
{o_{u1},\ldots,o_{uT_u}}$, and the label-sequence for utterance $u$
by $W_u$, the MMI criterion can be defined by the formula
\begin{equation}
F_{MMI} = \sum_u \log \frac{p(O_u|S_u)^{\alpha}p(W_u)}{\sum_W
p(O_u|S)^{\alpha}p(W)}, \label{eq_mmi_1}
\end{equation}
\noindent where $S_u = {s_{u1}, \ldots, s_{uT_u}}$ is the sequence
of states corresponding to $W_u$, and $\alpha$ is the acoustic
scaling factor. The sum in the denominator is taken over all phoneme
sequences in the decoded speech lattice for $u$. Differentiating
Eq.~(\ref{eq_mmi_1}) with respect to the log-likelihood $\log
p(o_{ut}|r)$ for state $r$ at time $t$, we get
\begin{align} \label{eq_mmi}
\frac{\partial F_{MMI}}{\partial \log p(o_{ut}|r)} &= \alpha\delta_{r;s_{ut}} - \frac{\alpha\sum_{W:s_t=r}p(O_u|S)^{\alpha}p(W)}{\sum_{W}p(O_u|S)^{\alpha}p(W)} \\
&= \alpha\big(\delta_{r;s_{ut}} - \gamma_{ut}^{DEN}(r)\big), \notag
\end{align}

\noindent where $\gamma_{ut}^{DEN}(r)$ is the posterior probability
of being in state $r$ at time $t$, computed over the denominator
lattices for utterance $u$ using the forward-backward algorithm, and
$\delta_{r;s_{ut}}$ is the Kronecker delta function (the binary
frame-level phonetic targets).

\section{Performing DNN Flat Start with MMI}
\label{section_mmiflatstart}

Sequence training criteria like the MMI error function are now
widely used in DNN training. However, all authors initialize their
networks using CE training, and apply the sequence-discriminative
criterion only in the final phase of the training procedure, to
fine-tune their models~\cite{Vesely-str, wiesler2015investigations},
which makes it necessary to use some method (HMM/GMM or iterative CE
training) to provide frame-level state targets. In contrast with
these authors, here we propose to apply MMI training in the flat
start phase. In order to be able to perform flat start of randomly
initialized DNNs using sequence training, we made some slight
changes in the standard MMI process, which we will describe next.

Firstly, we use the numerator occupancies $\gamma_{ut}^{NUM}(r)$ in
Eq.~(\ref{eq_mmi}) instead of the $\delta_{r;s_{ut}}$ values. This
way we can work with smoother targets instead of the crude binary
ones usually employed during DNN training. Another advantage of
eliminating the $\delta_{r;s_{ut}}$ values is that it allows us to
skip the preceding (usually GMM-based) label alignment step,
responsible for generating the frame-level training targets. We
applied the forward-backward algorithm to obtain the
$\gamma_{ut}^{NUM}(r)$ values, which solution has been mentioned in
some studies (e.g.~\cite{Vesely-str,Dong-seq}), but we only found
Zhou et al.~\cite{zhou2014sequence} actually doing this. However,
they pre-trained their DNN with the CE criterion first, while we
apply MMI training from the beginning, starting with randomly
initialized weights.


The second difference is that sequence training is conventionally
applied only to refine a fully trained system. Thus, the MMI
training criterion is calculated with CD phone models and a
word-level language model. This makes the decoding process slow, and
hence the numerator and denominator lattices are calculated only
once, before starting MMI training. In contrast to this, we execute
sequence DNN training using only phone-level transcripts and CI
phone models. This allows very fast decoding, so we can recalculate
the lattices after each sentence. This difference is crucial for the
fast convergence of our procedure. For converting the orthographic
transcripts to phone sequences one can follow the strategy of HTK.
That is, in the very first step we get the phonetic transcripts from
the dictionary, with no silences between the words. Pronunciation
alternatives and the optional short pause at word endings can be
added later on, performing realignment with a sufficiently
well-trained model~\cite{htkbook}.

A further difference is that we use no state priors or language
model, which makes the $\alpha$ scaling factor in Eq.~(\ref{eq_mmi})
unnecessary as well. Next, to reduce the computational requirements
of the algorithm, we estimated $\gamma_{ut}^{DEN}(r)$ using just the
most probable decoded path instead of summing over all possible
paths in the lattice (denoted by $\hat{\gamma}_{ut}^{DEN}(r)$).

With these modifications, the gradient with respect to the output
activations ($a_{ut}$) of the DNN is found using
\begin{align}
\frac{\partial F_{MMI}}{\partial a_{ut}(s)} &= \sum_{r}
\frac{\partial F_{MMI}}{\partial \log p(o_{ut}|r)} \frac{\partial
\log p(o_{ut}|r)}{\partial a_{ut}(s)} \\
&= \gamma_{ut}^{NUM}(s)-\hat{\gamma}_{ut}^{DEN}(s), \notag
\end{align}
which can be applied directly for DNN training. A standard technique
in DNN training is to separate a hold-out set from the training data
(see e.g.~\cite{rennie2014annealed}). If the error increases on this
hold-out set after a training iteration, then the DNN weights are
restored from a backup and the training continues with a smaller
learning rate. This strategy can be readily adapted to sequence DNN
training~\cite{kingsbury2009latticebased}, and we found it to be
essential for the stability of our flat-start MMI DNN training
method.
\begin{table}
\begin{framed}
\begin{itemize}
\item[(1)]{Frame-level phonetic targets ($\gamma_{ut}^{NUM}(r)$) are determined by a forward-backward search.}
\item[(2)]{We use only phoneme-level transcripts and CI phoneme states.}
\item[(3)]{We do not use state priors or language model.}
\item[(4)]{We estimate $\gamma_{ut}^{DEN}(r)$ by just the most probable decoded path ($\hat{\gamma}_{ut}^{DEN}(r)$).}
\item[(5)]{We measure training error on a hold-out set; when the error increases after a training iteration, we restore the weights and lower the learning rate.}
\end{itemize}
\end{framed} \vspace{-4mm}\caption{{\it Summary of our modifications on MMI training for DNN flat start.}} \label{table_mmi_modifications}
\end{table}

Table~\ref{table_mmi_modifications} summarizes the modifications
that we made to make MMI suitable for DNN flat start. Note that
steps (1) through (4) seek to simplify the procedure both to speed
it up and to make it more robust. Step (2) also helps us to perform
sequence-discriminative DNN training before CD state tying, which is
essential for applying it in flat start. Step (5), however, is
applied in our general DNN training process, but we found it
essential to avoid the ``runaway silence model'' issue which is a
common side effect haunting sequence-discriminative DNN training.

\section{KL divergence-based CD state tying}

%
Having aligned the CI phone models using flat-start training, the
next step is the construction of CD models. Currently, the dominant
solution for this is the decision tree-based state tying
method~\cite{Young1994treebased}. This technique pools all context
variants of a state, and then builds a decision tree by successively
splitting this set into two, according to one of the pre-defined
questions. For each step, it fits Gaussians on the distribution of
the states, and chooses the question which leads to the highest
likelihood gain. However, modeling the distribution of states with a
Gaussian function might be suboptimal when we utilize DNNs in the
final acoustic model.

To this end, we decided to first train an auxiliary neural network
on the CI target labels and then perform the CD state tying based on
the output of this network. Such a frame-level output can be treated
as a discrete probability distribution, and a natural distance
function for such distributions is the Kullback-Leibler (KL)
divergence~\cite{kullbackleibler}. Therefore, to control the state
tying process, we utilized the KL divergence-based decision
criterion introduced by Imseng et
al.~\cite{idiap2012feb,imseng2012comparing}. We basically replaced
the Gaussian-based likelihood function with a KL-divergence based
state divergence function; in other respects, the mechanism of the
CD state tying process remained the same. With this technique we
were not only able to eliminate GMMs from the state tying process,
but we also achieved a $4\%$ reduction in WER. For details,
see~\cite{gosztolya2015building}.

\section{Experimental Setup}

Our experimental setup is essentially the same as that of our
previous study~\cite{gosztolya2015building}.
We employed a DNN with 5 hidden layers, each containing 1000
rectified neurons~\cite{Glorot}, while the softmax activation
function was applied in the output layer. We used our custom DNN
implementation for GPU, which achieved outstanding results on
several datasets
(e.g.~\cite{Toth2014interspeech,grosz2015assessing,gosztolya2014detectingtheintensity,toth2015automatic,kovacs2015jointoptimization}).
We used 40 mel filter bank energies as features along with their
first and second order derivatives. Decoding and evaluation was
performed by a modified version of HTK~\cite{htkbook}.

The 28 hour-long speech corpus of Hungarian broadcast
news~\cite{grosz2013acomparison} was collected from eight TV
channels. The training set was about 22 hours long, a small part (2
hours) was used for validation purposes, and a 4-hour part was used
for testing.
We used a trigram language model and a vocabulary of 500k word
forms.
The order of utterances was randomized at the beginning of training.
We configured the state tying algorithms to get roughly 600, 1200,
1800, 2400, 3000 and 3600 tied states.

%
We tested four approaches for flat start training (i.e. to get the
frame-level phonetic targets for CD state tying and CE DNN
training). Firstly, we applied the standard GMM-based flat-start
training to produce initial time-aligned labels. To further improve
the segmentation, we trained a shallow CI ANN using the CE criterion
and re-aligned the frame labels based on the outputs of this ANN (we
will refer to this approach as the {\it ``GMM + ANN''} method). (In
our previous study we found that using a deep network for this
re-alignment setup did not give any significant
improvement~\cite{gosztolya2015building}.) After the realignment, we
applied both the standard GMM-based and our KL-criterion algorithms
for state tying. Then KL-based state tying was performed on the
output of the CI ANN.

%
Besides the standard GMM flat start approach, we evaluated the two
algorithms presented in sections \ref{section_flatstart} and
\ref{section_mmiflatstart} for flat starting with DNNs. In these
tests we always used five-hidden-layer CI DNNs. For the flat-start
method with iterative CE training ({\it ``Iterative CE''}) we
performed four training-aligning iterations, and KL-based CD state
tying was performed using the output and the alignments created by
the final DNN. For MMI training ({\it ``MMI''}) we also commenced
with a randomly initialized CI DNN. After applying the
discriminative sequence training method proposed in
Section~\ref{section_mmiflatstart}, the resulting DNN was used to
create forced aligned labels and also to provide the input posterior
estimates for KL clustering. In the last flat start approach tested,
we first applied the sequence-discriminative flat start method (i.e.
``MMI''). Then, based on the alignments of this network, we trained
another DNN with the CE criterion to supply both the final frame
labels and the likelihoods for KL-based CE state tying ({\it ``MMI +
CE''}).
%
%

The aim of this study was to compare various flat-start strategies.
This is why, after obtaining the CD labels, the final DNN models
were trained starting from randomly initialized weights and using
just the CE criterion. Of course, it might be possible to extend the
training with a final refinement
step using sequence-discriminative training.

\begin{table}[!t]
\centerline{\renewcommand{\arraystretch}{1.2}
\begin{tabular}{|l|l||c|c|c|}
\hline Flat start & State tying & \multicolumn{2}{c|}{WER \%} & No. of \\
\cline{3-4}
method & method & Dev. & Test & epochs \\
\hline\hline
GMM + ANN & GMM &~18.83\%~&~17.27\%~&--- \\
\hline
GMM + ANN & KL &~17.12\%~&~16.54\%~&--- \\
\hline\hline
Iterative CE &\multirow{3}{*}{KL} &~16.81\%~&~16.50\%~&~48~ \\
\cline{1-1}\cline{3-5}
MMI &&~16.50\%~&~15.96\%~&~13~ \\
\cline{1-1}\cline{3-5}
MMI + CE &&~16.36\%~&~15.86\%~&~29~ \\
\hline
\end{tabular}}
\caption{\label{table_all} {\it Word error rates (WER) for the
different flat start and state tying strategies.}}
\end{table}
\begin{figure*}[!t]
\begin{minipage}[b]{1.0\linewidth}
  \centering
  \centerline{\includegraphics[width=8.5cm]{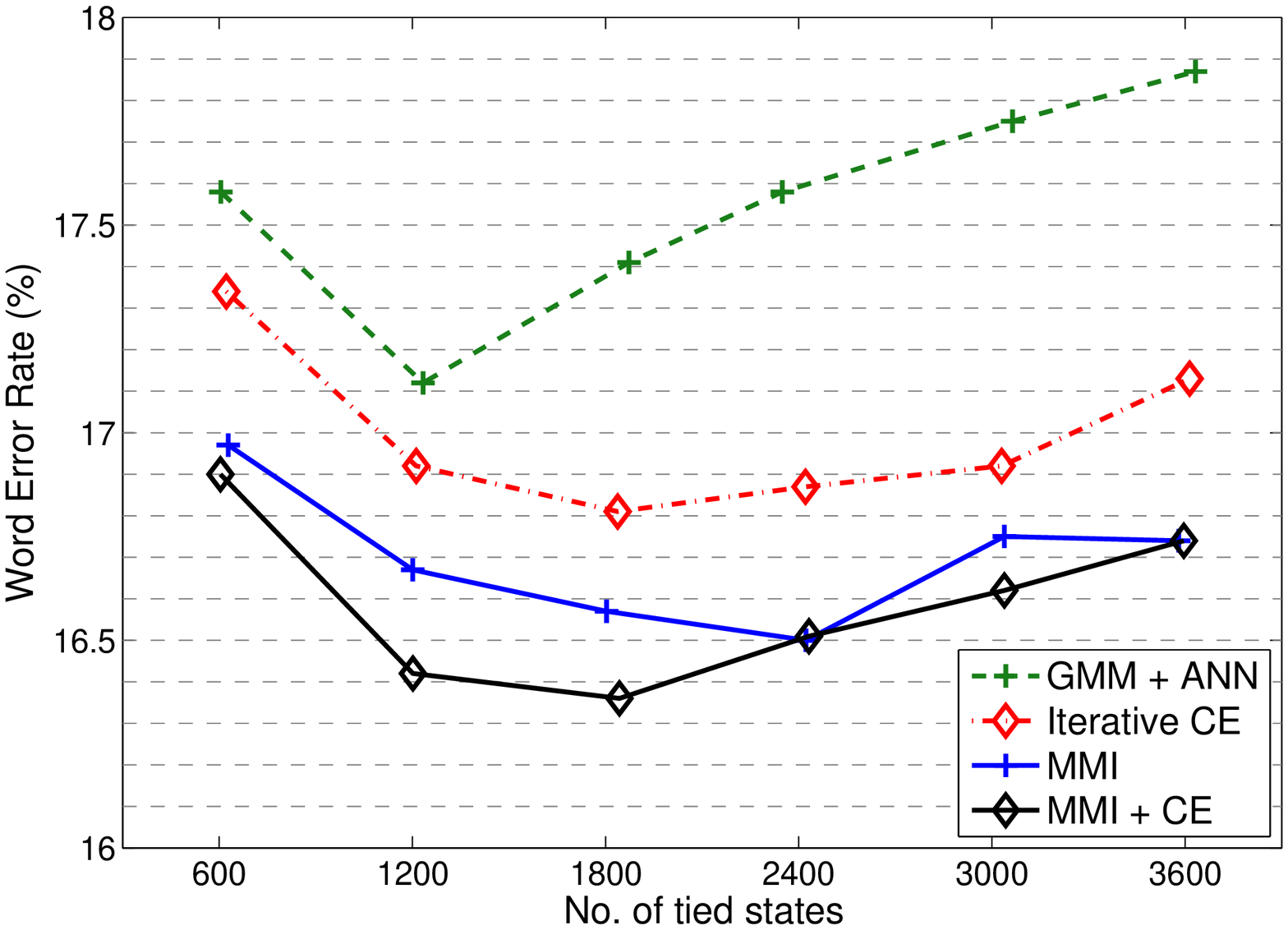}\hspace{0.5cm}\includegraphics[width=8.5cm]{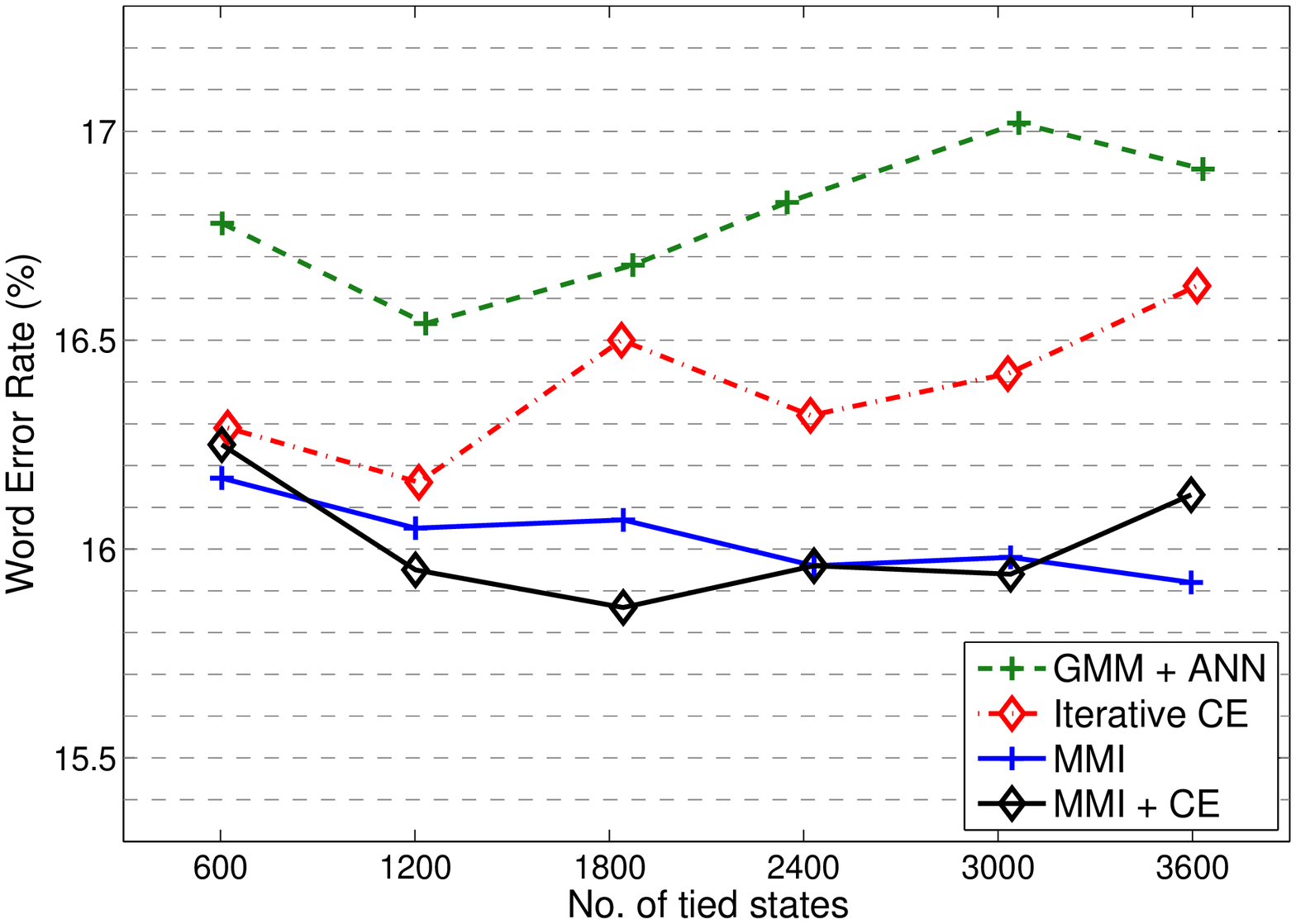}}
\caption{WER as a function of the number of KL-clustered tied states
on the development (left) and test (right) sets.} \label{fig_res}
\end{minipage}
\end{figure*}

\section{Results and Discussion}
\label{sec_results}

Figure~\ref{fig_res} shows the resulting WER scores as a function of
the number of CD tied states. As can be seen, the MMI-based flat
start strategy gave slightly better results than the iterative
method in every case. We also observed that the final CD models
which got their training labels from the MMI-trained DNN were more
stable with respect to varying the number of CD states. Fine-tuning
the labels of the MMI-trained DNN with a CE-trained DNN (``MMI'' vs.
``MMI+CE'') seems unnecessary, as it was not able to significantly
improve the results. This indicates that sequence training yields
both fine alignments and good posterior
estimates.

Table~\ref{table_all} summarizes the best WER values on the
development set, and the corresponding scores on the test set. The
KL clustering method clearly outperformed the GMM-based state tying
technique. Comparing the alignment methods, we see that relying on
the alignments produced by the HMM/GMM resulted in the lowest
accuracy score, in spite of the fine-tuning step using an ANN. With
the parameter configurations applied, the iterative CE training
method performed slightly worse than the MMI-based strategies.
Unfortunately, for the iterative CE method the right number of
training-aligning steps is hard to tune.
For example, Zhang et al. performed 20 such
iterations~\cite{zhang2014standalone}, while we employed only 4
iterations. In this respect, it is more informative to compare the
training times
, which are shown in the rightmost column of Table~\ref{table_all}.
(We did not indicate the number of epochs for the ``GMM + ANN''
method, as the training procedure was radically different there.)
For our 28-hour dataset, 48 epochs were required by the four
iterations of iterative CE flat-start training, while MMI required
only one-fourth of it; and although performing the forward-backward
search adds a slight overhead to the training process, it is clear
that MMI was still much faster, even when the final CE re-alignment
step was also involved.

Measuring the training times in 
CPU/GPU time gives even larger differences in favor of the MMI
method (3 hours vs. 16 hours). The reason is that for iterative CE
flat-start training we used a mini-batch of 100 frames (which we
found optimal previously~\cite{gosztolya2015building}), while for
MMI whole utterances (usually more than 1000 frames) were used to
update the weights, and this allowed better parallelization on the
GPU.

In our view, two modifications are crucial for the speed and
stability of the proposed algorithm. The first one is that we use
only CI phone models without phone language model, so we can very
quickly update the numerator and denominator lattices after the
processing of each sentence.
This continuous refinement of the frame-level soft targets obviously
leads to a faster convergence. The only study we know of, which does
not perform the re-alignment of the frame-level targets strictly
after a training iteration, is that of Bacchiani et
al.~\cite{bacchiani2014asynchronous}. Their study focuses on
describing their massively parallelized online neural network
optimization system, where a separate thread is responsible for the
alignment of the phonetic targets, while DNN training is performed
by the client machines. Besides the fact that in their model there
is no guarantee for that the alignment of phonetic targets are
up-to-date, it is easy to see that their architecture is quite
different from a standard DNN training architecture, making their
techniques pretty hard to adapt. In contrast, our slight
modifications can be applied relatively easily.

As regards stability, a known drawback of sequence training methods
is that the same process is responsible both for aligning and
training the DNN, which often leads to the ``run-away silence
model'' issue~\cite{Hang-mmi-sil}. That is, after a few iterations,
only one model (usually the silence model) dominates most parts of
the utterances, which is even reinforced in the next training step.
To detect the occurrence of this phenomenon, we monitored the error
rate on a hold-out set during training. If the error increased after
an iteration, we restored the weights of the network to their
previous values and the learning rate was halved. In our experience,
restoring the weights to their previous values and continuing the
training using a lower learning rate can successfully handle this
issue.


\section{Conclusions}

Here, we showed how to perform flat start with
sequence-discriminative DNN training. We applied the standard MMI
sequence training method, for which we introduced several minor
modifications. Our results showed that, compared to the standard
procedure of iterative CE DNN training and re-alignment, not only
were we able to reduce the WER scores, but we also achieved a
significant reduction in training times. By also utilizing the
Kullback-Leibler divergence-based CD state tying method proposed
earlier, we made the whole training procedure of context-dependent
HMM/DNNs GMM-free.

%

\eightpt
\bibliographystyle{IEEEtran}
\bibliography{2016-interspeech-mmi}
\end{document}